\documentclass[11pt]{article}

\usepackage[final]{acl}

\usepackage{times}
\usepackage{latexsym}

\usepackage[T1]{fontenc}

\usepackage[utf8]{inputenc}

\usepackage{microtype}

\usepackage{inconsolata}

\usepackage{graphicx}

\usepackage{booktabs}
\usepackage{tabularx}
\usepackage[table]{xcolor}
\usepackage{multirow}
\usepackage{algorithm}
\usepackage{algorithmic}
\usepackage{amsmath}   
\usepackage{amsfonts}  
\usepackage{amssymb}  
\usepackage{enumitem}
\usepackage{mdframed}
\usepackage{float}
\usepackage{xcolor}
\usepackage{booktabs}
\usepackage{multirow}

\usepackage{times}
\usepackage{soul}
\usepackage{url}
\usepackage{amsthm}
\usepackage{booktabs}
\usepackage{algorithm}
\usepackage{algorithmic}
\usepackage[switch]{lineno}       
\usepackage[T1]{fontenc}         
\usepackage{amsfonts}       
\usepackage{nicefrac}       
\usepackage{microtype}      
\usepackage{xcolor}         
\usepackage{tikz}                         
\usepackage{subcaption}    
\usepackage{apxproof}            
\usepackage{wrapfig}       
\usepackage{multirow}       
\usepackage{pifont}         
\usepackage{mathtools}                 
\usepackage{comment,bbm}    
\usepackage{textcomp}       
\usepackage{xspace}  

\theoremstyle{plain}

\theoremstyle{definition}

\theoremstyle{remark}

\theoremstyle{takeaways}
\newtheorem{takeaways}{Takeaways}

\title{Self-Awareness before Action: Mitigating Logical Inertia via Proactive Cognitive Awareness}

\author{
Fulong Fan\textsuperscript{1,\textdagger},
Peilin Liu\textsuperscript{1,\textdagger},
Liu FengZhe\textsuperscript{1},
Shuyan Yang\textsuperscript{2},
Gang Yan\textsuperscript{2,*} \\
\textsuperscript{1}School of Software, Jilin University \\
\textsuperscript{2}School of Computer Science, Jilin University \\
\textsuperscript{\textdagger}Equal contribution \quad
\textsuperscript{*}Corresponding author \\
\texttt{\{fanfl9922, liupl9922, liufz9922, yangsy9922\}@mails.jlu.edu.cn} \\
\texttt{gyan8@jlu.edu.cn}
}

\begin{document}
\maketitle

\begin{abstract}

Large language models perform well on many reasoning tasks, yet they often lack awareness of whether their current knowledge or reasoning state is complete. In non-interactive puzzle settings, the narrative is fixed and the underlying structure is hidden; once a model forms an early hypothesis under incomplete premises, it can propagate that error throughout the reasoning process, leading to unstable conclusions. To address this issue, we propose SABA, a reasoning framework that explicitly introduces self-awareness of missing premises before making the final decision. SABA formulates reasoning as a recursive process that alternates between structured state construction and obstacle resolution: it first applies Information Fusion to consolidate the narrative into a verifiable base state, and then uses Query-driven Structured Reasoning to identify and resolve missing or underspecified premises by turning them into queries and progressively completing the reasoning state through hypothesis construction and state refinement. Across multiple evaluation metrics, SABA achieves the best performance on all three difficulty splits of the non-interactive Detective Puzzle benchmark, and it also maintains leading results on multiple public benchmarks.
\end{abstract}

\section{Introduction}
\label{sec:intro}

Large language models have shown strong ability in multi-step reasoning and narrative understanding. 
In interactive settings such as social games, agents can acquire new information through dialogue and revise their beliefs over time~\cite{
zhang2024exploringcollaborationmechanismsllm,
song2025survivalevaluatingllmssocial, wang2023avalonsgamethoughtsbattle, zhu2025playerenhancingllmbasedmultiagent, wu-etal-2024-deciphering}. 
In contrast, in non-interactive puzzle settings, the narrative is fixed and no new information can be obtained. 
The model must recover the hidden truth only from a long text that contains implicit clues, missing links, and distractors. 
In this regime, the main challenge is not how to ask for information, but how to locate, align, and connect the information that already exists in the narrative~\cite{liu2023lostmiddlelanguagemodels}. 
A small early mistake can remain uncorrected and can guide all later reasoning, which often leads to unstable and incorrect conclusions~\cite{10.1145/3571730,dziri2023faithfatelimitstransformers}. 
Existing work on abductive and long-context reasoning reports that current models still struggle under this form of information asymmetry~\cite{del2023truedetectivedeepabductive,piekos-etal-2021-measuring,wang2024understandingreasoningabilitylanguage}. 
This observation motivates the need for a reasoning process that can correct such early errors.

\begin{figure}[t]
	\centering
	\includegraphics[width=0.95\columnwidth]{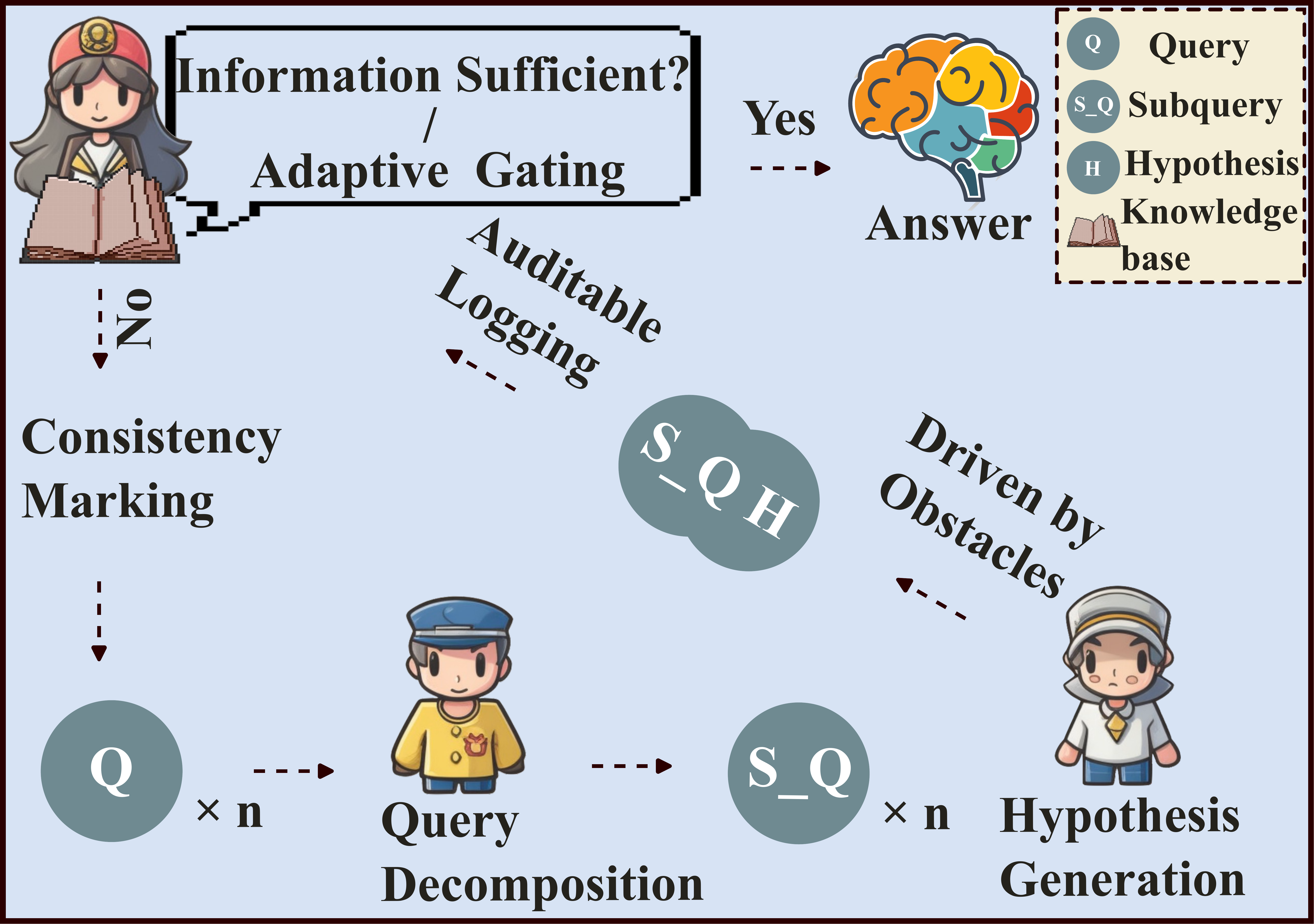} 
    \vspace{-0.1in}
	\caption{Overview of the SABA Framework.}
    \vspace{-0.15in}
	\label{fig:framework_overview}
\end{figure}

Most existing reasoning paradigms are not well suited to this setting. 
Prompt-based methods such as chain-of-thought tend to commit to an early hypothesis and then expand it, even when the initial premise is weak~\cite{10.5555/3600270.3602070,turpin2023languagemodelsdontsay}.
Decomposition methods introduce intermediate steps, but they often lose global coherence when the narrative is long and the evidence is scattered~\cite{zhou2023leasttomost,khot2023decomposedpromptingmodularapproach}. 
Refinement-based methods revise an answer after it is produced, but they often justify the same early mistake instead of triggering a full re-evaluation, which leads to confirmation bias~\cite{huang2024largelanguagemodelsselfcorrect,stechly2023gpt4doesntknowits}. 
These limitations suggest that a reliable agent should not start from answering the task, but should first examine whether its current understanding is complete and consistent~\cite{shinn2023reflexionlanguageagentsverbal}. 
This perspective shifts the focus from direct prediction to state assessment.

Based on this view, we propose SABA, a reasoning framework that treats the model as a system that audits its own knowledge state before making a decision. 
As shown in Figure~\ref{fig:framework_overview}, the core idea of SABA is to perform long-horizon reasoning through a recursive control loop that alternates between structured state construction and obstacle-driven reasoning. 
SABA consists of two modules. The first module is \textbf{Information Fusion (IF)}, which transforms the raw narrative into a structured and verified baseline by aligning events with attributes and by annotating logical consistency. 
This step reduces dispersion, weak signals, and hidden relations in the text. 
The second module is \textbf{Query-driven Structured Reasoning (QSR)}, which treats missing or unclear premises as explicit obstacles. 
These obstacles are converted into queries and are resolved through hypothesis construction and state enrichment. 
This process continues until the state is sufficient to support a final conclusion. 
This modular structure supports both clarity and control during inference.

This design allows the model to make missing premises and latent causal gaps explicit, and to resolve them before committing to an answer. 
As a result, SABA reduces logical leaps, limits unsupported assumptions, and supports stable reasoning over long narratives. 
The reasoning process is also transparent, since each step updates an explicit state and records what information was added and why. 
This property supports later inspection and analysis.
We evaluate SABA on a non-interactive detective benchmark called Detective Puzzle and on several general reasoning benchmarks. 
On the most difficult split of Detective Puzzle, SABA achieves a clear improvement over strong baselines in both answer correctness and evidence grounding. 

Overall, our contributions are as follows. 
First, we focus on a form of non-interactive narrative reasoning where the main difficulty is truth reconstruction under information asymmetry. 
Second, we propose SABA, a framework that performs awareness before action through explicit state construction and obstacle resolution. 
Third, we introduce an evaluation setting that measures both final correctness and evidence use, which supports a more detailed analysis of reasoning behavior.

\section{Related Work}
Our work relates to three directions for improving complex reasoning in large language models, addressing a specific under-explored failure mode: in non-interactive long narratives, models often commit early based on incomplete premises and propagate this error.

\subsection{The Challenge of Long-Context Narrative Reasoning}
Unlike interactive settings where agents can query for missing information~\cite{zhu2025playerenhancingllmbasedmultiagent,wu-etal-2024-deciphering}, non-interactive narratives present fixed, diffuse evidence with distractors. This amplifies the lost-in-the-middle effect, causing early retrieval errors to compound~\cite{dziri2023faithfatelimitstransformers,del2023truedetectivedeepabductive}. Decomposition strategies (e.g., Least-to-Most~\cite{zhou2023leasttomost}) reduce local difficulty but do not ensure global coherence when clues are interdependent and scattered. The bottleneck is aligning dispersed evidence into a consistent, reusable situation model. SABA addresses this via an \textit{Information Fusion} module, which consolidates the narrative into a unified, verifiable event-centric state before final deduction, making latent associations explicit and preventing evidence dilution.

\subsection{Limitations of Post-Hoc Refinement}
Methods like Self-Refine~\cite{madaan2023selfrefineiterativerefinementselffeedback} and Reflexion~\cite{shinn2023reflexionlanguageagentsverbal} iteratively refine an initial output using self-feedback. However, such answer-then-correct paradigms are prone to confirmation bias~\cite{huang2024largelanguagemodelsselfcorrect,stechly2023gpt4doesntknowits}: subsequent steps rationalize the initial conclusion rather than re-audit its premises. This is severe in non-interactive narratives where a flawed early hypothesis dominates. SABA avoids this by shifting refinement from the candidate answer to the underlying knowledge state, enforcing an explicit audit for completeness and consistency before commitment.

\subsection{From Unstructured Chains to Structured State Management}
Works like Chain-of-Thought~\cite{10.5555/3600270.3602070} and Tree-of-Thoughts~\cite{yao2023treethoughtsdeliberateproblem} externalize reasoning traces but operate over unstructured text, lacking explicit representation of missing or inconsistent information. Recent advances in structured state tracking~\cite{wang2024understandingreasoningabilitylanguage} suggest explicit state management improves reliability, yet state completion under missing premises remains open. SABA formalizes reasoning as iterative construction and verification of a structured state. It introduces \textit{Query-driven Structured Reasoning} to treat missing/conflicting premises as active obstacles, transforming them into targeted queries for state repair—converting narrative reasoning into a deliberate process of filling a verified knowledge state.

\noindent \textbf{Summary.} Prior work separately addresses long-context challenges, iterative refinement, and structured traces. SABA synthesizes these into a state-first framework that builds and audits a coherent knowledge foundation before answer generation, mitigating premature commitment in non-interactive narrative reasoning.

\section{Proposed Method}

The core idea of SABA is to perform long-horizon narrative reasoning via an explicit recursive control loop that alternates between (i) structured information fusion and consistency annotation, and (ii) obstacle-driven query decomposition and hypothesis generation. This design explicitly represents missing premises and latent causal gaps, thereby reducing logical leaps and information forgetting in standard chain-of-thought reasoning.
This recursive structure allows the system to repeatedly refine its internal state, which supports stable reasoning over long and complex narratives.

\subsection{Information Fusion}

Let $D=\{x_1,\dots,x_n\}$ denote the raw narrative units. The goal of Information Fusion (IF) is to transform $D$ into a structured and verified baseline state for subsequent reasoning.
IF explicitly organizes and annotates narrative evidence to counteract dispersion, weak signaling, and implicit relations in long texts, which otherwise hinder reliable retrieval and reuse.
This transformation ensures that later reasoning stages operate on an explicit and stable representation rather than on fragmented textual traces.

\paragraph{Event Alignment.}
We decompose $D$ into a backbone sequence of core events
\begin{equation}
S = \{s_1,\dots,s_m\},
\end{equation}
and a set of heterogeneous attributes
\begin{equation}
A = \{a_1,\dots,a_p\},
\end{equation}
including actions, object states, locations, and evidentiary descriptors.

This decomposition separates the \emph{narrative skeleton} (what happened and in what order) from the \emph{descriptive details} (what properties, objects, or side conditions are involved), which are often scattered and weakly localized in the text.
The alignment step aims to explicitly bind these descriptive attributes back to the events they qualify, thereby making implicit associations explicit and retrievable.
This explicit binding reduces the need for later inference steps to reconstruct such links from memory alone.
We then define an alignment mapping:
\begin{equation}
\Phi_{\text{map}}: A \rightarrow 2^S,
\end{equation}
which assigns each attribute to one or more backbone events. This yields aligned units
\begin{equation}
d_i = \big(s_i,\{a \in A \mid s_i \in \Phi_{\text{map}}(a)\}\big),
\end{equation}
and the aligned sequence $D_{\text{aligned}}=\{d_1,\dots,d_m\}$.

Operationally, $\Phi_{\text{map}}$ is implemented by jointly considering semantic relevance, temporal proximity, and entity overlap between $a$ and $s$.
Allowing a multi-assignment ($A \to 2^S$) ensures that attributes that span multiple events (e.g., persistent object states or long-term intentions) are not artificially localized to a single point, which would otherwise reintroduce information loss.
This design choice preserves long-range dependencies that are critical for coherent narrative interpretation.

\paragraph{Consistency Check.}
While alignment densifies information, it does not guarantee that the resulting structure is logically coherent.
We therefore perform an explicit consistency annotation step to detect potential conflicts and uncertainties that may later mislead the reasoning process.
This step complements alignment by adding a layer of logical validation.

For each aligned unit $d_i$, we compute a verification comment:
\begin{equation}
b_i = \psi_{\text{vfy}}(d_i, D_{\text{aligned}}\setminus d_i),
\end{equation}
where $\psi_{\text{vfy}}$ checks temporal, entity-state, and causal consistency. The result is the baseline state:
\begin{equation}
D_{\text{base}} = \{(d_i,b_i)\}_{i=1}^m.
\end{equation}

Each comment $b_i$ serves as a localized diagnostic signal indicating whether $d_i$ is potentially problematic, underspecified, or contradictory relative to the rest of the narrative.
These annotations are not used to discard information, but to mark it with explicit uncertainty or risk, which can then be explicitly handled by later reasoning stages instead of being silently ignored.
This approach ensures that uncertainty remains visible and actionable throughout the reasoning process.

\paragraph{Empirical Analysis.}
\begin{figure}[t]
    \centering
    \includegraphics[width=0.98\columnwidth]{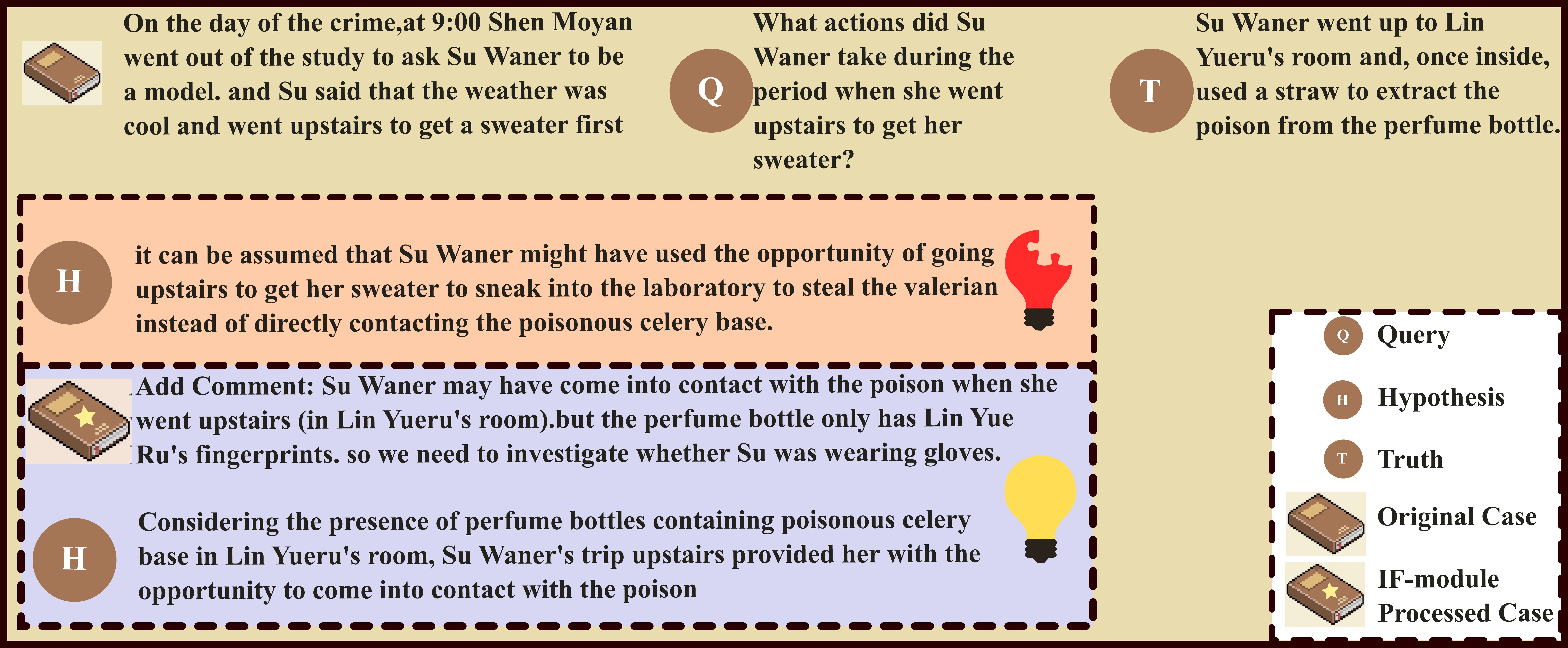}
    \vspace{-0.1in}
    \caption{Visualization of the IF module's impact. The pre-association of dispersed clues provides a verified cognitive baseline for subsequent reasoning.}
    \vspace{-0.15in}
    \label{fig:IF}
\end{figure}

As illustrated in Figure~\ref{fig:IF}, standard LLMs often suffer from the ``lost-in-the-middle'' effect and clue dilution within long narratives. While traditional models may identify explicit individual clues, they frequently overlook implicit associations, for instance, failing to connect the action of ``going upstairs'' with the presence of ``poison storage.'' 
The IF module establishes a verified cognitive baseline by pre-associating such dispersed attributes.
By annotating latent risks (e.g., ``potential contact with poison''), IF ensures that critical evidence remains highly available throughout the reasoning trajectory, effectively preventing the information forgetting prevalent in vanilla CoT.
This empirical observation supports the functional role of IF in stabilizing long-horizon reasoning.

\begin{takeaways}
Information Fusion converts an unstructured narrative into a dense, structured evidence representation.
Feature Alignment makes implicit associations explicit and retrievable across reasoning steps.
Consistency Annotation externalizes uncertainty and conflict, preventing silent propagation of errors.
\end{takeaways}

\subsection{Query-driven Structured Reasoning}

Given a task $T$ (e.g., determining Motive, Modus Operandi, and Final Judgment), SABA performs recursive reasoning over the state $D_t$.
The key idea of Query-driven Structured Reasoning (QSR) is to explicitly expose missing premises, resolve them via targeted queries, and incrementally enrich the reasoning state until a logically sufficient basis for $T$ is reached.
This process treats reasoning as a progressive construction of support rather than as a single inference step.

\paragraph{Gap Identification.}
Instead of directly attempting to infer $T$, the model first diagnoses what is missing.
At iteration $t$, it identifies a set of reasoning obstacles
\begin{equation}
\Omega_t = \mathcal{M}(p_{\text{aware}} \mid D_t, T),
\end{equation}
where each obstacle $\omega$ represents a missing or underspecified premise required to infer $T$ from $D_t$.
This step ensures that reasoning failures are addressed at their source rather than after they propagate.

Each obstacle is normalized as
\begin{equation}
\omega = (\tau(\omega), \mathrm{dim}(\omega), \mathrm{req}(\omega)),
\end{equation}
where $\tau(\omega)$ is the type (e.g., MissingLink, Ambiguity, MotiveGap), $\mathrm{dim}(\omega)$ is the blocked task dimension, and $\mathrm{req}(\omega)$ is the missing requirement.
This representation makes gaps in the reasoning process explicit and manipulable.
It also enables systematic handling of different forms of incompleteness.

Each obstacle $\omega_i$ is then decomposed into sub-queries
\begin{equation}
Q_{i,t} = \mathcal{M}(p_{\text{dec}} \mid \omega_i, D_t),
\end{equation}
such that each $q \in Q_{i,t}$ targets a specific component of $\mathrm{req}(\omega_i)$.
This decomposition converts abstract reasoning gaps into concrete information needs that can be individually addressed.
As a result, each gap becomes a well-defined unit of further analysis.

\paragraph{Hypothesis Construction.}
For each sub-query $q$, a hypothesis is generated as
\begin{equation}
h = \mathcal{M}(p_{\text{hypo}} \mid q, D_t),
\end{equation}
yielding $H_{i,t}=\{h\}$.
These hypotheses act as tentative logical bridges that fill the detected gaps and are later validated through consistency and interaction with other evidence.
They allow the system to explore possible explanations in a controlled manner.
The reasoning state is updated as:
\begin{equation}
D_{t+1} = D_t \cup Q_t \cup H_t,
\end{equation}
where $Q_t=\bigcup_i Q_{i,t}$ and $H_t=\bigcup_i H_{i,t}$.
This update accumulates resolved premises while preserving the full trace of how each piece of information was introduced.
Such traceability supports both transparency and later error analysis.

The recursion terminates when either $\Omega_t=\emptyset$ (logical closure) or a maximum depth $t_{\max}$ is reached. The final conclusion is synthesized as
\begin{equation}
y = \mathcal{M}(p_{\text{syn}} \mid D_t, T).
\end{equation}
This termination rule ensures that the process is both complete and computationally bounded.

\paragraph{Empirical Analysis.}
\begin{figure}[t]
    \centering
    \includegraphics[width=0.98\columnwidth]{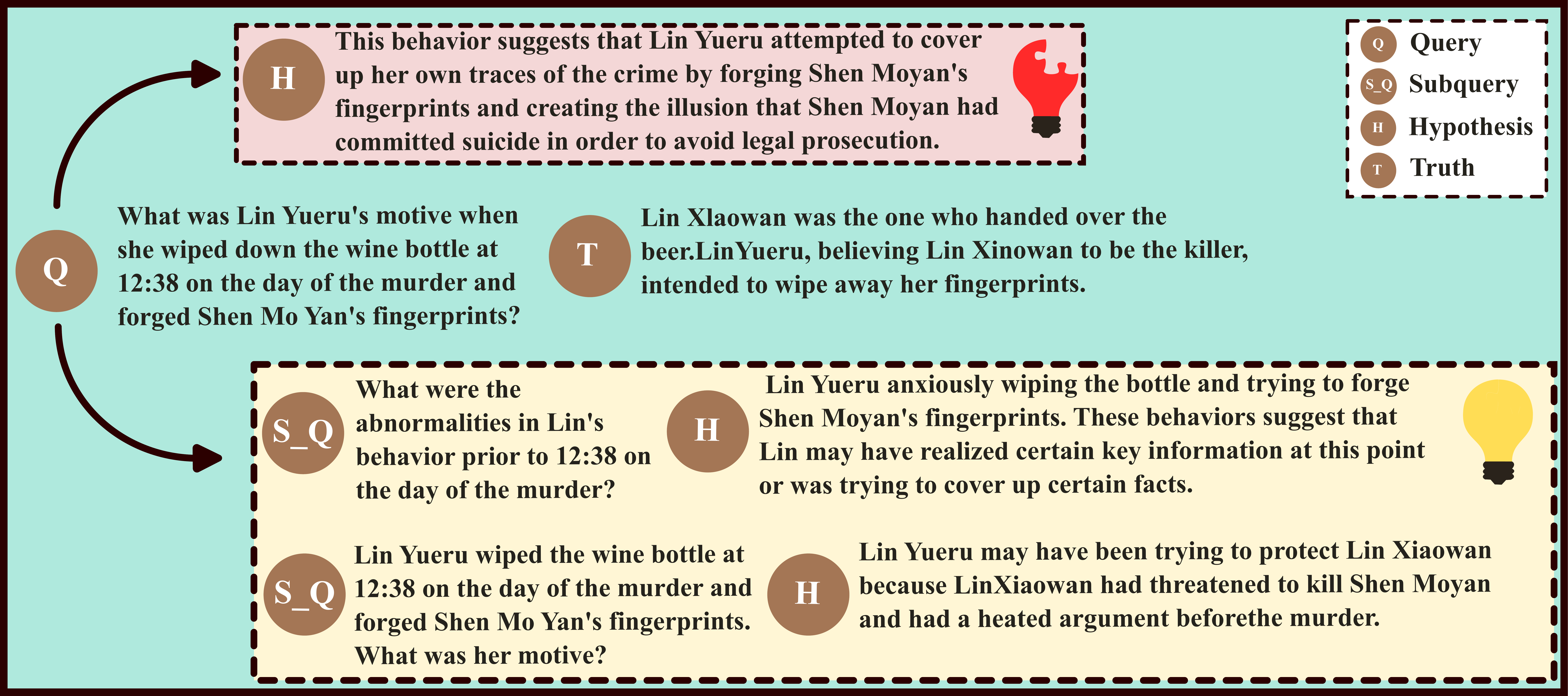}
    \vspace{-0.1in}
    \caption{Visualization of the QSR module's impact: How query decomposition resolves logical obstacles and identifies hidden evidence.}
    \vspace{-0.15in}
    \label{fig:QSR}
\end{figure}

As shown in Figure~\ref{fig:QSR}, traditional CoT often incurs ``logical leaps'', leading to premature accusations based on superficial proximity. In contrast, QSR treats the absence of a causal link as a formal reasoning obstacle. Through recursive decomposition, it generates heuristic hypotheses ($H$) that successfully uncover the deep-seated motivation behind specific actions (e.g., ``protecting Lin Xiaowan'').
This behavior illustrates how QSR reduces unsupported inference while increasing causal depth.

\begin{takeaways}
QSR reframes reasoning as iterative gap detection and resolution rather than direct deduction.
Obstacle decomposition converts abstract missing premises into concrete, answerable queries.
Recursive enrichment yields a transparent and logically grounded reasoning trajectory.
\end{takeaways}

\subsection{Workflow of SABA}
Algorithm~1 summarizes the complete SABA reasoning procedure. Given a raw narrative $D$ and a target task $T$, the algorithm first constructs a structured and verified baseline via Information Fusion (Phase~1), and then iteratively refines this baseline via Query-driven Structured Reasoning (Phase~2). To optimize efficiency, an adaptive gating mechanism is employed between phases: the system evaluates the density of logical \textit{Conflicts} (alignment contradictions) and \textit{Doubts} (missing causal links) within $D_{base}$. If both indicators fall below predefined thresholds, the iterative loop is bypassed in favor of immediate synthesis, preventing redundant computation on straightforward tasks.The core idea is to treat reasoning as progressive state completion: instead of directly inferring the answer, the model incrementally constructs the set of premises required to support a valid conclusion.

Specifically, the algorithm alternates between (i) diagnosing missing premises (obstacles), (ii) decomposing them into concrete queries, and (iii) generating hypotheses that fill these gaps. 
This loop continues until no critical obstacles remain or a maximum recursion depth is reached, at which point final synthesis is performed.
This iterative structure ensures that reasoning proceeds in a controlled and verifiable manner.

\begin{algorithm}[!ht]
\caption{SABA Recursive Reasoning}
\small
\begin{mdframed}[backgroundcolor=gray!10, linewidth=0pt, innerleftmargin=5pt, innerrightmargin=5pt, innertopmargin=5pt, innerbottommargin=5pt]
\begin{algorithmic}[1]
\REQUIRE Raw narrative $D$, Task $T$, Max depth $t_{max}$, Thresholds $x, y$
\ENSURE Final conclusion $y$

\STATE \COMMENT{\textbf{Phase 1: Information Fusion (IF)}}
\STATE $D_{aligned} \leftarrow \mathcal{M}(p_{align} \mid D)$ \COMMENT{Feature Alignment}
\STATE $D_{base} \leftarrow \{ (d_i, \mathcal{M}(p_{vfy} \mid d_i)) \mid d_i \in D_{aligned} \}$
\STATE $\mathbb{C}, \mathbb{D} \leftarrow \text{Extract\_Metrics}(D_{base})$ \COMMENT{Conflicts and Doubts}

\STATE \vspace{0.3em}
\STATE \COMMENT{\textbf{Adaptive Gating}}
\IF{$|\mathbb{C}| \le x$ \AND $|\mathbb{D}| \le y$}
    \STATE \textbf{goto} \textbf{Final Synthesis}
\ENDIF

\STATE \vspace{0.3em}
\STATE \COMMENT{\textbf{Phase 2: Query-driven Structured Reasoning (QSR)}}
\STATE $D_0 \leftarrow D_{base}, t \leftarrow 0$
\WHILE{$t < t_{max}$}
    \STATE $\Omega_t \leftarrow \mathcal{M}(p_{aware} \mid D_t, T)$ \COMMENT{Obstacle Identification}
    \IF{$\Omega_t = \emptyset$}
        \STATE \textbf{break} \COMMENT{Logical Closure}
    \ENDIF
    
    \STATE $Q_t, H_t \leftarrow \emptyset, \emptyset$
    \FOR{\textbf{each} obstacle $\omega_i \in \Omega_t$}
        \STATE $Q_{i,t} \leftarrow \mathcal{M}(p_{dec} \mid \omega_i, D_t)$ \COMMENT{Query Decomposition}
        \STATE $H_{i,t} \leftarrow \{ \mathcal{M}(p_{hypo} \mid q, D_t) \mid \forall q \in Q_{i,t} \}$ \COMMENT{Hypothesis Generation}
        \STATE $Q_t \leftarrow Q_t \cup Q_{i,t}, H_t \leftarrow H_t \cup H_{i,t}$
    \ENDFOR
    
    \STATE $D_{t+1} \leftarrow D_t \cup \{Q_t, H_t\}$ \COMMENT{State Enrichment}
    \STATE $t \leftarrow t + 1$
\ENDWHILE

\STATE \vspace{0.3em}
\STATE \COMMENT{\textbf{Final Synthesis}} \label{alg:saba:final}
\STATE $y \leftarrow \mathcal{M}(p_{syn} \mid D_{t}, T)$ \COMMENT{Final Deduction}
\RETURN $y$
\end{algorithmic}
\end{mdframed}
\end{algorithm}

At each iteration $t$, the model logs the state
\begin{equation}
\mathcal{L}_t = (\Omega_t, Q_t, H_t, D_t),
\end{equation}
which explicitly records what was missing, what was queried, what was hypothesized, and how the reasoning state evolved. This trace enables post-hoc inspection, error analysis, and verification of logical consistency.
It also supports reproducibility by making the reasoning path observable.

Overall, the workflow ensures that information is introduced only when required to resolve a specific obstacle, which limits unnecessary speculation and constrains the propagation of unsupported assumptions. By transforming implicit chains of thought into explicit state transitions, SABA produces a transparent and auditable reasoning process.
This design improves both reliability and interpretability in long-horizon narrative reasoning.

\section{Experiments}
\label{sec:experiments}

\subsection{Experimental Setup}

\paragraph{Benchmarks.}
\label{subsec:benchmarks}
To evaluate SABA under varying degrees of information asymmetry, we adopt a complementary set of reasoning benchmarks. Dataset statistics are summarized in Table~\ref{tab:dataset-stats}.

(1) Detective Puzzle (DP). Our primary benchmark for non-interactive endogenous truth reconstruction. It contains 31 cases with three difficulty levels: \textit{Easy} (direct inference), \textit{Medium} (multi-step causality), and \textit{Complex} (implicit clues with red herrings). In our collection, 28 cases are sourced from the ``5-Minute Mystery'' platform, and the remaining 3 cases are adapted from classic detective novels.

(2) Multi-hop Reasoning. We include HotpotQA~\cite{yang2018hotpotqadatasetdiverseexplainable} to test evidence integration across disconnected passages, and StrategyQA~\cite{geva2021didaristotleuselaptop} to assess bridging implicit commonsense gaps.

(3) Big-Bench Hard (BBH). We report the average accuracy over 23 challenging tasks from BBH~\cite{suzgun2022challengingbigbenchtaskschainofthought} to verify whether SABA generalizes beyond detective-style deduction.

\begin{table}[t]
\centering
\footnotesize
\setlength{\tabcolsep}{3.5pt}
\renewcommand{\arraystretch}{1.08}
\caption{Benchmark statistics.}
\label{tab:dataset-stats}
\resizebox{\columnwidth}{!}{%
\begin{tabular}{lccc}
\toprule
\textbf{Category} & \textbf{Level} & \textbf{Samples} & \textbf{Avg. Length} \\
\midrule
\multirow{3}{*}{\textbf{Detective Puzzle}} & Easy    & 5  & 1050 words \\
                                           & Medium  & 15 & 1150 words \\
                                           & Complex & 11 & 950 words \\
\midrule
\textbf{General} & \multicolumn{3}{l}{HotpotQA, StrategyQA, Big-Bench Hard} \\
\bottomrule
\end{tabular}%
}
\end{table}

\paragraph{Evaluation Metrics.}
\label{subsec:metrics}
We evaluate whether a model reconstructs the truth via a coherent evidentiary chain rather than relying on superficial cues. For DP, we report both correctness and evidence-grounding metrics; for general benchmarks, we follow standard protocols.
Specifically, for DP we report Suspect Accuracy (SA) for perpetrator identification, along with Motive Recall (R-M), Modus Operandi Recall (R-O), and Clue Coverage Rate (CCR). CCR measures the breadth of critical clue exploration.

To measure evidence grounding for motive and modus operandi, we compute semantic recall via atomic semantic matching: we decompose both prediction and reference into atomic propositions, compute embedding similarity, and apply greedy one-to-one matching with threshold $\tau=0.5$.

For general benchmarks, we follow standard protocols. For HotpotQA, we report Answer EM (Ans) and Supporting Facts F1 (SF). For StrategyQA and BBH, we report accuracy.

\paragraph{Baselines.}
\label{subsec:baselines}
We compare SABA against three categories of prompting paradigms.

(1) Basic Reasoning: Direct prompting, Chain-of-Thought (CoT)~\cite{10.5555/3600270.3602070}, and Self-Consistency (SC)~\cite{wang2023selfconsistencyimproveschainthought}.

(2) Decomposition and Planning: SELF-DISCOVER~\cite{zhou2024selfdiscoverlargelanguagemodels}, and Graph-of-Thought (GoT)~\cite{Besta_2024}.

(3) Iterative Optimization: Self-Refine~\cite{madaan2023selfrefineiterativerefinementselffeedback}, CRITIC~\cite{gou2024criticlargelanguagemodels}, and $S^2R$~\cite{ma2025s2rteachingllmsselfverify}.

\paragraph{Implementation Details.}

We implement SABA using DeepSeek-V3 and Gemini-1.5-Flash as backbone models. To improve reproducibility and reduce decoding randomness, we set the decoding temperature to 0.0 for all experiments and use default API settings for other parameters. For semantic similarity computation, we use all-MiniLM-L6-v2~\cite{reimers-gurevych-2019-sentence}. Unless otherwise specified, all performance metrics are reported as mean $\pm$ standard deviation over three independent runs under identical experimental settings, while the normalized inference cost $T$ is reported as the average relative cost across the same three runs, with Direct set to 1.0.

\subsection{Experimental Results}
\label{subsubsec:Experimental Results and Analysis}

\begin{table*}[t]
\setlength{\extrarowheight}{0.6pt}
\centering
\small
\setlength{\tabcolsep}{5.4pt}
\renewcommand{\arraystretch}{1.14}
\renewcommand\multirowsetup{\centering}

\caption{Main results of \textbf{DeepSeek-V3} across DP and general benchmarks. For DP, each cell reports results for three difficulty levels (\textit{Easy} / \textit{Medium} / \textit{Complex}) from top to bottom. All reported results, except the normalized inference cost $T$, are shown as mean $\pm$ standard deviation over three independent runs. \textbf{HQA} denotes HotpotQA, \textbf{SQA} denotes StrategyQA. All results are shown in \% except for $T$.}
\label{tab:ds_results}

\begin{tabular*}{\textwidth}{@{\extracolsep{\fill}} ccccc|cc|c|c|c @{}}
\toprule
\multirow{2}{*}{\textbf{\normalsize Method}}
& \multicolumn{4}{c}{\textbf{DP}}
& \multicolumn{2}{c}{\textbf{HQA}}
& \multicolumn{1}{c}{\textbf{SQA}}
& \multicolumn{1}{c}{\textbf{BBH}}
& \multirow{2}{*}{\textbf{$T$}} \\
\cmidrule(lr){2-5}
\cmidrule(lr){6-7}
\cmidrule(lr){8-8}
\cmidrule(lr){9-9}
& \multicolumn{1}{c}{\textbf{SA}}
& \multicolumn{1}{c}{\textbf{R-M}}
& \multicolumn{1}{c}{\textbf{R-O}}
& \multicolumn{1}{c}{\textbf{CCR}}
& \multicolumn{1}{c}{\textbf{Ans}}
& \multicolumn{1}{c}{\textbf{SF}}
& \multicolumn{1}{c}{\textbf{Acc}}
& \multicolumn{1}{c}{\textbf{Acc}}
& \multicolumn{1}{c@{}}{} \\
\midrule

\multirow{3}{*}{\normalsize Direct}
& 65.2$\pm$0.5 & 65.0$\pm$0.7 & 60.5$\pm$0.8 & 71.0$\pm$0.7
& \multirow{3}{*}{62.4$\pm$0.5}
& \multirow{3}{*}{52.3$\pm$0.6}
& \multirow{3}{*}{82.0$\pm$0.4}
& \multirow{3}{*}{78.7$\pm$0.5}
& \multirow{3}{*}{\textbf{1.0}} \\
& 48.1$\pm$0.8 & 64.3$\pm$0.8 & 58.0$\pm$0.9 & 62.5$\pm$0.9 & & & & & \\
& 40.7$\pm$0.9 & 59.4$\pm$1.0 & 58.9$\pm$1.0 & 58.7$\pm$1.0 & & & & & \\
\midrule

\multirow{3}{*}{\normalsize CoT}
& 68.1$\pm$0.7 & 77.5$\pm$0.9 & 65.1$\pm$0.9 & 75.9$\pm$0.8
& \multirow{3}{*}{68.4$\pm$0.7}
& \multirow{3}{*}{60.3$\pm$0.7}
& \multirow{3}{*}{87.6$\pm$0.5}
& \multirow{3}{*}{86.0$\pm$0.6}
& \multirow{3}{*}{2.5} \\
& 58.6$\pm$1.0 & 64.9$\pm$1.0 & 62.0$\pm$1.0 & 72.2$\pm$1.0 & & & & & \\
& 45.4$\pm$1.1 & 61.0$\pm$1.1 & 59.7$\pm$1.1 & 61.9$\pm$1.2 & & & & & \\
\midrule

\multirow{3}{*}{\normalsize Self-Refine}
& 70.9$\pm$1.0 & 66.9$\pm$1.1 & 68.5$\pm$1.2 & 80.3$\pm$1.0
& \multirow{3}{*}{68.6$\pm$0.8}
& \multirow{3}{*}{59.0$\pm$0.9}
& \multirow{3}{*}{88.5$\pm$0.7}
& \multirow{3}{*}{87.3$\pm$0.8}
& \multirow{3}{*}{6.3} \\
& 65.5$\pm$1.2 & 66.7$\pm$1.2 & 65.0$\pm$1.2 & 76.4$\pm$1.2 & & & & & \\
& 55.6$\pm$1.3 & 62.0$\pm$1.4 & 61.4$\pm$1.4 & 65.5$\pm$1.5 & & & & & \\
\midrule

\multirow{3}{*}{\normalsize SC($k$=5)}
& 72.0$\pm$1.2 & 68.8$\pm$1.2 & 69.2$\pm$1.3 & 81.5$\pm$1.2
& \multirow{3}{*}{72.0$\pm$0.9}
& \multirow{3}{*}{62.4$\pm$1.0}
& \multirow{3}{*}{90.1$\pm$0.8}
& \multirow{3}{*}{89.6$\pm$0.9}
& \multirow{3}{*}{12.0} \\
& 70.3$\pm$1.3 & 64.6$\pm$1.4 & 66.4$\pm$1.4 & 79.9$\pm$1.3 & & & & & \\
& 62.7$\pm$1.5 & 62.1$\pm$1.5 & 62.3$\pm$1.6 & 71.3$\pm$1.6 & & & & & \\
\midrule

\multirow{3}{*}{\normalsize CRITIC}
& 77.5$\pm$1.0 & 67.9$\pm$1.1 & 70.8$\pm$1.1 & 88.6$\pm$1.0
& \multirow{3}{*}{74.0$\pm$0.8}
& \multirow{3}{*}{68.7$\pm$0.9}
& \multirow{3}{*}{90.4$\pm$0.7}
& \multirow{3}{*}{88.0$\pm$0.8}
& \multirow{3}{*}{29.3} \\
& 73.6$\pm$1.1 & 65.6$\pm$1.2 & 68.5$\pm$1.2 & 87.6$\pm$1.1 & & & & & \\
& 66.4$\pm$1.3 & 64.6$\pm$1.3 & 63.5$\pm$1.4 & 73.6$\pm$1.4 & & & & & \\
\midrule

\multirow{3}{*}{\normalsize $S^2R$}
& 79.3$\pm$1.1 & 70.8$\pm$1.0 & 76.3$\pm$1.3 & 89.7$\pm$1.1
& \multirow{3}{*}{73.4$\pm$0.8}
& \multirow{3}{*}{68.7$\pm$0.9}
& \multirow{3}{*}{91.7$\pm$0.7}
& \multirow{3}{*}{90.7$\pm$0.8}
& \multirow{3}{*}{18.4} \\
& 76.3$\pm$1.2 & 66.5$\pm$1.3 & 72.2$\pm$1.4 & 88.0$\pm$1.3 & & & & & \\
& 68.4$\pm$1.4 & 63.7$\pm$1.4 & 67.2$\pm$1.6 & 77.1$\pm$1.4 & & & & & \\
\midrule

\multirow{3}{*}{\normalsize SELF-DISC.}
& 81.0$\pm$1.0 & 72.3$\pm$1.1 & 78.5$\pm$1.1 & 87.1$\pm$1.0
& \multirow{3}{*}{75.6$\pm$0.7}
& \multirow{3}{*}{68.3$\pm$0.9}
& \multirow{3}{*}{92.5$\pm$0.7}
& \multirow{3}{*}{91.0$\pm$0.8}
& \multirow{3}{*}{5.0} \\
& 75.1$\pm$1.2 & 68.4$\pm$1.2 & 69.9$\pm$1.3 & 86.9$\pm$1.1 & & & & & \\
& 68.8$\pm$1.3 & 64.2$\pm$1.4 & 65.3$\pm$1.4 & 75.6$\pm$1.4 & & & & & \\
\midrule

\multirow{3}{*}{\normalsize GoT}
& 84.1$\pm$1.2 & 78.6$\pm$1.3 & 82.2$\pm$1.1 & 90.5$\pm$1.2
& \multirow{3}{*}{78.0$\pm$0.8}
& \multirow{3}{*}{73.2$\pm$1.0}
& \multirow{3}{*}{91.7$\pm$0.8}
& \multirow{3}{*}{90.7$\pm$0.9}
& \multirow{3}{*}{35.7} \\
& 77.0$\pm$1.4 & 79.9$\pm$1.4 & 72.1$\pm$1.2 & 87.9$\pm$1.4 & & & & & \\
& 69.8$\pm$1.6 & 69.8$\pm$1.2 & 66.9$\pm$1.7 & 77.3$\pm$1.7 & & & & & \\
\midrule

\multirow{3}{*}{\normalsize SABA}
& \textbf{85.7$\pm$0.7} & \textbf{84.3$\pm$0.9} & \textbf{85.8$\pm$1.3} & \textbf{93.8$\pm$0.6}
& \multirow{3}{*}{\textbf{78.6$\pm$0.7}}
& \multirow{3}{*}{\textbf{73.5$\pm$0.8}}
& \multirow{3}{*}{\textbf{94.4$\pm$0.4}}
& \multirow{3}{*}{\textbf{93.2$\pm$0.5}}
& \multirow{3}{*}{9.2} \\
& \textbf{83.2$\pm$1.1} & \textbf{74.9$\pm$0.7} & \textbf{73.5$\pm$0.6} & \textbf{92.2$\pm$0.9} & & & & & \\
& \textbf{79.3$\pm$1.2} & \textbf{73.4$\pm$0.8} & \textbf{72.7$\pm$1.4} & \textbf{83.3$\pm$0.6} & & & & & \\
\bottomrule
\end{tabular*}
\end{table*}

\begin{table*}[t]
\centering
\scriptsize
\setlength{\tabcolsep}{4pt}
\renewcommand{\arraystretch}{1.08}
\caption{Controlled ablation on DeepSeek-V3. Results are reported as mean $\pm$ standard deviation over three independent runs (\%).}
\label{tab:ablation}
\resizebox{\textwidth}{!}{
\begin{tabular}{lcccccc}
\toprule
Variant & DP-Complex (SA) & DP-Complex (CCR) & StrategyQA & HQA (Ans) & HQA (SF) & BBH \\
\midrule
SABA (Full)                & 79.3$\pm$1.2 & 83.3$\pm$0.6 & 94.4$\pm$0.4 & 78.6$\pm$0.7 & 73.5$\pm$0.8 & 93.2$\pm$0.5 \\
w/o IF                     & 69.8$\pm$1.1 & 70.7$\pm$0.9 & 82.2$\pm$0.6 & 70.7$\pm$0.6 & 63.7$\pm$1.0 & 87.4$\pm$0.7 \\
Self-assessment-only       & 65.8$\pm$1.3 & 65.9$\pm$1.1 & 79.1$\pm$0.8 & 67.1$\pm$0.8 & 59.8$\pm$1.1 & 82.5$\pm$0.9 \\
w/o Awareness              & 61.7$\pm$1.5 & 62.2$\pm$1.2 & 76.7$\pm$0.9 & 62.8$\pm$1.0 & 56.5$\pm$1.1 & 78.7$\pm$1.0 \\
\bottomrule
\end{tabular}
}
\end{table*}

Comprehensive evaluation results are detailed in Table~\ref{tab:ds_results} for DeepSeek-V3 and in \hyperref[app:gemini_results]{Appendix~\ref*{app:gemini_results}} (Table~\ref{tab:gemini_results}) for Gemini-1.5-Flash. Across both models, SABA consistently achieves the strongest overall performance among compared methods, particularly on long-context complex reasoning tasks.

\paragraph{Performance on Detective Puzzle Benchmarks.}
On DeepSeek-V3, SABA achieves SA scores of 85.7 $\pm$ 0.7, 83.2 $\pm$ 1.1, and 79.3 $\pm$ 1.2 on the Easy, Medium, and Complex splits, respectively. The largest gain is observed on the Complex split, where SABA improves SA from 69.8 $\pm$ 1.6 (the strongest baseline, GoT) to 79.3 $\pm$ 1.2, yielding a gain of 9.5 percentage points. Meanwhile, SABA also achieves the strongest evidence-grounding performance on Complex, reaching CCR 83.3 $\pm$ 0.6 (vs.\ 77.1 $\pm$ 1.4 for $S^2R$ and 77.3 $\pm$ 1.7 for GoT), R-M 73.4 $\pm$ 0.8 (vs.\ 69.8 $\pm$ 1.2 for GoT), and R-O 72.7 $\pm$ 1.4 (vs.\ 67.2 $\pm$ 1.6 for $S^2R$), indicating more thorough exploration of critical clues and more faithful reconstruction of motive and modus operandi under implicit gaps and red herrings.

In contrast, as task difficulty increases, linear prompting strategies such as CoT exhibit weaker evidence coverage on Complex (CCR 61.9$\pm$1.2).
Optimization-based methods such as Self-Refine improve SA on Complex (55.6$\pm$1.3) but remain limited in evidence grounding (CCR 65.5$\pm$1.5), suggesting that iterative refinement may still overlook missing premises when the narrative contains strong distractors.
Results on Gemini-1.5-Flash (Table~\ref{tab:gemini_results}) show a similar trend, demonstrating robustness across models of varying scales.

\paragraph{Generalization Across General Reasoning Benchmarks.}
To verify that SABA's advantage stems from general reasoning principles, we evaluated it on mainstream benchmarks.
As shown in Table~\ref{tab:ds_results} (right), SABA improves HotpotQA Answer EM to 78.6$\pm$0.7 and Supporting Facts F1 to 73.5$\pm$0.8, slightly surpassing the strongest baseline GoT (78.0$\pm$0.8 / 73.2$\pm$1.0).
On StrategyQA and BBH, SABA achieves accuracies of 94.4$\pm$0.4 and 93.2$\pm$0.5, respectively, maintaining a clear lead over the strongest baselines.
These results are consistent with the view that explicitly identifying and resolving missing premises can benefit both long-context narrative reasoning and multi-hop reasoning settings.

\paragraph{Inference Efficiency Analysis.}
As shown in Table~\ref{tab:ds_results}, SABA achieves a normalized inference cost of $T=9.2$, compared with $T=12.0$ for SC and $T=35.7$ for GoT. This means that SABA reduces inference cost by 23.3\% relative to SC and by 74.2\% relative to GoT, while still delivering consistently stronger reasoning performance. These results suggest that SABA allocates additional computation in a more targeted and effective manner rather than uniformly increasing overhead, thanks to its adaptive mechanism and caching strategy. In particular, the substantial gap from GoT indicates that SABA avoids the heavy cost often associated with more complex structured reasoning pipelines, while preserving the benefits of deeper causal analysis. Combined with the clear gains in SA and evidence-grounding metrics (R-M/R-O/CCR), the results show that the extra computation is effectively used to improve causal resolution and logical consistency, yielding a favorable trade-off between reasoning quality and inference cost for long-context complex reasoning tasks.

\paragraph{Controlled Ablation Analysis.}

To identify which component is responsible for SABA's overall gain, we conduct controlled ablations on DeepSeek-V3 and evaluate all variants on DP-Complex, StrategyQA, HQA, and BBH. For evaluation, we use SA and CCR on DP-Complex, Accuracy on StrategyQA and BBH, and Ans and SF on HQA. We compare four settings: (1) \textit{SABA (Full)}, which keeps the complete framework and serves as the reference model; (2) \textit{w/o IF}, which removes Information Fusion while retaining awareness and the reasoning loop, and is used to examine the contribution of consolidating scattered evidence into a grounded intermediate state; (3) \textit{Self-assessment-only}, which preserves gap identification but removes grounded state construction and controlled completion, and is used to test whether self-assessment alone is sufficient; and (4) \textit{w/o Awareness}, which further disables awareness-driven obstacle identification, and is used to examine the role of explicit obstacle diagnosis in preventing premature commitment under incomplete evidence.

As shown in Table~\ref{tab:ablation}, performance degrades consistently as more structure is removed, and the relative decline clearly reflects the role of each module. Removing Information Fusion already causes a clear drop, with DP-Complex SA decreasing from 79.3$\pm$1.2 to 69.8$\pm$1.1, corresponding to a relative decline of 12.0\%, and CCR decreasing from 83.3$\pm$0.6 to 70.7$\pm$0.9, corresponding to a relative decline of 15.1\%. This suggests that SABA benefits from first consolidating scattered clues into a grounded intermediate state before further reasoning. When the framework is reduced to \textit{Self-assessment-only}, performance drops further, indicating that gap awareness alone is not sufficient unless it is paired with grounded state construction and controlled completion. The largest degradation appears in \textit{w/o Awareness}, where DP-Complex SA falls from 79.3$\pm$1.2 to 61.7$\pm$1.5, corresponding to a relative decline of 22.2\%, and StrategyQA falls from 94.4$\pm$0.4 to 76.7$\pm$0.9, corresponding to a relative decline of 18.8\%. This suggests that explicit obstacle identification is the key mechanism for preventing premature commitment under incomplete evidence. Overall, the results show that SABA works best when awareness, evidence consolidation, and controlled reasoning-state completion operate together, rather than in isolation. To further assess the reliability of hypothesis bridging in QSR, we provide an additional analysis in \hyperref[app:hypothesis_reliability]{Appendix~\ref*{app:hypothesis_reliability}}, including the proportions of unsupported, flagged, and subsequently corrected hypotheses.

\paragraph{Model Generality and Dependence.}
\begin{table}[t]
\centering
\small
\caption{SABA performance with Llama-3.1-70B on DP benchmarks.}
\label{tab:llama_results}
\begin{tabular}{l cccc}
\toprule
\textbf{Level} & \textbf{SA} & \textbf{R-M} & \textbf{R-O} & \textbf{CCR} \\
\midrule
Easy   & 82.5$\pm$1.2 & 67.3$\pm$0.8 & 73.8$\pm$1.5 & 84.5$\pm$1.5 \\
Medium & 73.3$\pm$1.3 & 69.8$\pm$1.1 & 67.4$\pm$0.8 & 80.6$\pm$1.3 \\
Complex & 68.1$\pm$0.8 & 68.9$\pm$0.9 & 63.2$\pm$1.6 & 75.8$\pm$1.8 \\
\bottomrule
\end{tabular}
\end{table}

To evaluate SABA's robustness across different architectures, we conducted an additional study using Llama-3.1-70B. As shown in Table~\ref{tab:llama_results}, SABA delivers stable performance across all difficulty levels, achieving a CCR of $84.5 \pm 1.5$ on \textit{Easy}, $80.6 \pm 1.3$ on \textit{Medium}, and $75.8 \pm 1.8$ on \textit{Complex}. These results indicate that the framework maintains relatively stable evidence-grounding ability even when instantiated on a backbone different from DeepSeek-V3.

At the same time, the absolute performance remains lower than that of DeepSeek-V3, especially on the more challenging \textit{Complex} split. This suggests that the effectiveness of SABA is still influenced by the underlying model's base reasoning capacity. In other words, SABA does not eliminate backbone differences, but provides a structured reasoning process that can be transferred across architectures and continue to function effectively under different model families.

Overall, the Llama-3.1-70B results provide additional evidence that SABA is not tied to a single backbone. Its gains do not appear to come solely from a particular model-specific prompting effect; rather, the framework itself contributes a portable reasoning structure that remains effective across architectures. Meanwhile, the remaining performance gap also indicates that stronger backbone competence can further improve the quality of obstacle identification, evidence integration, and final decision making within the same framework.

\section{Conclusions and Limitations}
\label{sec:conclusion}

In this paper, we introduced \textbf{SABA}, a reasoning paradigm based on the \textit{Self-Awareness Before Action} principle that decouples cognitive assessment from answer generation and frames deduction as an iterative process of gap identification and resolution.
This design treats reasoning as a controlled and stepwise process and aims to improve stability and reliability in complex tasks.

However, our study has several limitations. First, SABA depends on the self-evaluation ability of the backbone model, which can limit obstacle detection quality for smaller models. Second, its recursive process introduces higher latency, which may affect real-time use and practical deployment. Additionally, the framework relies on the IF module for structured input processing, and a fully end-to-end extraction of clues remains an open problem that is important for future improvement.

In future work, the integration of \textit{Retrieval-Augmented Generation (RAG)} could further improve factual grounding and reliability by providing external support for evidence and reducing residual errors caused by missing knowledge.

\section*{Acknowledgments}
This work was supported by the Excellent Young Scientists Fund (Overseas) of the National Natural Science Foundation of China, and in part by the Jilin Provincial Department of Science and Technology (Grant No.~20260102309JC). We also thank NODI Lab. for valuable support and insightful discussions.



\bibliography{custom}

\appendix

\section{Additional Experimental Results and Analyses}
\label{sec:appendix_results}

\subsection{Additional Results on Gemini-1.5-Flash}
\label{app:gemini_results}

To complement the main results on DeepSeek-V3, we further evaluate SABA on Gemini-1.5-Flash using the same benchmark suite and evaluation protocol. Specifically, we report SA, R-M, R-O, and CCR on Detective Puzzle (DP), Answer EM and Supporting Facts F1 on HotpotQA (HQA), and accuracy on StrategyQA (SQA) and BBH. Following the revised experimental setting in the main paper, all reported results, except the normalized inference cost $T$, are shown as mean $\pm$ standard deviation over three independent runs.

\begin{table*}[t]
\setlength{\extrarowheight}{0.6pt}
\centering
\small
\setlength{\tabcolsep}{5.4pt}
\renewcommand{\arraystretch}{1.14}
\renewcommand\multirowsetup{\centering}
\caption{Main results of \textbf{Gemini-1.5-Flash} across the three Detective Puzzle difficulty levels and general reasoning benchmarks. For DP, each cell reports results for \textit{Easy}, \textit{Medium}, and \textit{Complex} from top to bottom. All reported results, except the normalized inference cost $T$, are shown as mean $\pm$ standard deviation over three independent runs. \textbf{HQA} denotes HotpotQA, \textbf{SQA} denotes StrategyQA. All results are shown in \% except for $T$.}
\label{tab:gemini_results}

\begin{tabular*}{\textwidth}{@{\extracolsep{\fill}} ccccc|cc|c|c|c @{}}
\toprule
\multirow{2}{*}{\textbf{\normalsize Method}}
& \multicolumn{4}{c}{\textbf{DP}}
& \multicolumn{2}{c}{\textbf{HQA}}
& \multicolumn{1}{c}{\textbf{SQA}}
& \multicolumn{1}{c}{\textbf{BBH}}
& \multirow{2}{*}{\textbf{$T$}} \\
\cmidrule(lr){2-5}
\cmidrule(lr){6-7}
\cmidrule(lr){8-8}
\cmidrule(lr){9-9}
& \multicolumn{1}{c}{\textbf{SA}}
& \multicolumn{1}{c}{\textbf{R-M}}
& \multicolumn{1}{c}{\textbf{R-O}}
& \multicolumn{1}{c}{\textbf{CCR}}
& \multicolumn{1}{c}{\textbf{Ans}}
& \multicolumn{1}{c}{\textbf{SF}}
& \multicolumn{1}{c}{\textbf{Acc}}
& \multicolumn{1}{c}{\textbf{Acc}}
& \multicolumn{1}{c@{}}{} \\
\midrule

\multirow{3}{*}{\normalsize Direct}
& 40.1$\pm$0.8 & 48.2$\pm$0.9 & 47.1$\pm$1.1 & 45.3$\pm$1.2
& \multirow{3}{*}{52.2$\pm$0.8}
& \multirow{3}{*}{42.3$\pm$0.9}
& \multirow{3}{*}{72.1$\pm$0.7}
& \multirow{3}{*}{68.7$\pm$0.8}
& \multirow{3}{*}{\textbf{1.0}} \\
& 35.6$\pm$0.9 & 44.5$\pm$1.1 & 45.3$\pm$1.2 & 35.1$\pm$1.4 & & & & & \\
& 32.3$\pm$0.9 & 42.4$\pm$1.2 & 38.0$\pm$1.4 & 32.4$\pm$1.5 & & & & & \\
\midrule

\multirow{3}{*}{\normalsize CoT}
& 42.8$\pm$0.7 & 52.0$\pm$0.9 & 51.3$\pm$1.0 & 49.5$\pm$1.0
& \multirow{3}{*}{58.3$\pm$0.8}
& \multirow{3}{*}{48.2$\pm$0.9}
& \multirow{3}{*}{78.3$\pm$0.8}
& \multirow{3}{*}{75.0$\pm$0.9}
& \multirow{3}{*}{2.4} \\
& 38.0$\pm$1.0 & 45.9$\pm$1.1 & 48.0$\pm$1.1 & 48.1$\pm$1.1 & & & & & \\
& 35.7$\pm$1.2 & 44.8$\pm$1.2 & 40.7$\pm$1.3 & 35.5$\pm$1.4 & & & & & \\
\midrule

\multirow{3}{*}{\normalsize Self-Refine}
& 41.8$\pm$0.9 & 50.5$\pm$1.0 & 50.1$\pm$1.1 & 52.0$\pm$1.2
& \multirow{3}{*}{55.0$\pm$0.9}
& \multirow{3}{*}{46.7$\pm$1.0}
& \multirow{3}{*}{79.1$\pm$0.8}
& \multirow{3}{*}{76.3$\pm$0.9}
& \multirow{3}{*}{6.2} \\
& 38.5$\pm$1.0 & 48.2$\pm$1.1 & 44.0$\pm$1.2 & 42.4$\pm$1.3 & & & & & \\
& 32.7$\pm$1.3 & 45.4$\pm$1.3 & 43.2$\pm$1.4 & 38.2$\pm$1.5 & & & & & \\
\midrule

\multirow{3}{*}{\normalsize SC($k$=5)}
& 51.9$\pm$0.7 & 48.3$\pm$1.0 & 50.6$\pm$1.0 & 65.1$\pm$1.2
& \multirow{3}{*}{60.4$\pm$0.9}
& \multirow{3}{*}{49.9$\pm$1.0}
& \multirow{3}{*}{82.4$\pm$0.8}
& \multirow{3}{*}{79.5$\pm$0.9}
& \multirow{3}{*}{12.1} \\
& 51.2$\pm$1.2 & 45.1$\pm$1.1 & 45.0$\pm$1.2 & 58.7$\pm$1.3 & & & & & \\
& 42.5$\pm$1.4 & 42.7$\pm$1.3 & 42.4$\pm$1.4 & 45.2$\pm$1.6 & & & & & \\
\midrule

\multirow{3}{*}{\normalsize CRITIC}
& 58.0$\pm$1.0 & 52.4$\pm$1.0 & 65.0$\pm$1.1 & 72.4$\pm$1.1
& \multirow{3}{*}{62.1$\pm$0.9}
& \multirow{3}{*}{55.4$\pm$1.0}
& \multirow{3}{*}{68.1$\pm$0.8}
& \multirow{3}{*}{62.4$\pm$0.9}
& \multirow{3}{*}{29.2} \\
& 52.5$\pm$1.1 & 48.2$\pm$1.1 & 52.3$\pm$1.2 & 67.9$\pm$1.2 & & & & & \\
& 49.9$\pm$1.3 & 55.1$\pm$1.3 & 48.5$\pm$1.4 & 60.4$\pm$1.5 & & & & & \\
\midrule

\multirow{3}{*}{\normalsize $S^2R$}
& 65.7$\pm$1.1 & 58.0$\pm$1.0 & 67.5$\pm$1.1 & 74.8$\pm$1.2
& \multirow{3}{*}{62.7$\pm$0.9}
& \multirow{3}{*}{54.9$\pm$1.0}
& \multirow{3}{*}{83.1$\pm$0.8}
& \multirow{3}{*}{81.5$\pm$0.9}
& \multirow{3}{*}{18.5} \\
& 61.8$\pm$1.2 & 55.7$\pm$1.1 & 59.8$\pm$1.2 & 72.7$\pm$1.3 & & & & & \\
& 67.5$\pm$1.1 & 61.8$\pm$1.3 & 55.7$\pm$1.5 & 64.9$\pm$1.5 & & & & & \\
\midrule

\multirow{3}{*}{\normalsize SELF-DISC.}
& 63.0$\pm$0.9 & 56.7$\pm$1.0 & 69.0$\pm$1.1 & 74.5$\pm$1.2
& \multirow{3}{*}{65.0$\pm$0.8}
& \multirow{3}{*}{57.7$\pm$0.9}
& \multirow{3}{*}{85.0$\pm$0.8}
& \multirow{3}{*}{83.4$\pm$0.9}
& \multirow{3}{*}{4.8} \\
& 59.9$\pm$0.8 & 51.8$\pm$1.1 & 63.1$\pm$1.3 & 69.8$\pm$1.2 & & & & & \\
& 56.7$\pm$1.2 & 58.5$\pm$1.3 & 51.8$\pm$1.4 & 62.8$\pm$1.5 & & & & & \\
\midrule

\multirow{3}{*}{\normalsize GoT}
& 67.1$\pm$0.9 & 58.9$\pm$1.1 & 70.6$\pm$1.2 & 78.1$\pm$1.1
& \multirow{3}{*}{68.5$\pm$0.9}
& \multirow{3}{*}{60.9$\pm$1.0}
& \multirow{3}{*}{84.4$\pm$0.8}
& \multirow{3}{*}{82.7$\pm$0.9}
& \multirow{3}{*}{35.5} \\
& 64.0$\pm$1.0 & 56.4$\pm$1.2 & 62.1$\pm$1.1 & 75.5$\pm$1.3 & & & & & \\
& 61.0$\pm$1.2 & 62.1$\pm$1.4 & 58.5$\pm$1.5 & 65.1$\pm$1.6 & & & & & \\
\midrule

\multirow{3}{*}{\normalsize SABA}
& \textbf{74.3$\pm$0.8} & \textbf{65.0$\pm$0.9} & \textbf{72.7$\pm$1.1} & \textbf{85.4$\pm$1.0}
& \multirow{3}{*}{\textbf{70.3$\pm$0.8}}
& \multirow{3}{*}{\textbf{63.0$\pm$0.9}}
& \multirow{3}{*}{\textbf{87.4$\pm$0.7}}
& \multirow{3}{*}{\textbf{86.1$\pm$0.9}}
& \multirow{3}{*}{9.8} \\
& \textbf{72.2$\pm$1.0} & \textbf{68.8$\pm$1.0} & \textbf{64.9$\pm$1.2} & \textbf{82.9$\pm$1.1} & & & & & \\
& \textbf{68.8$\pm$1.1} & \textbf{72.1$\pm$1.3} & \textbf{62.4$\pm$1.4} & \textbf{75.7$\pm$1.5} & & & & & \\
\bottomrule
\end{tabular*}
\end{table*}

As a supplementary backbone study, Table~\ref{tab:gemini_results} reports the results of Gemini-1.5-Flash under the same benchmark suite and evaluation protocol as the main DeepSeek-V3 experiments. We evaluate all methods on the three difficulty splits of Detective Puzzle, namely Easy, Medium, and Complex, as well as on the general reasoning benchmarks HotpotQA, StrategyQA, and BBH. For DP, we report SA, R-M, R-O, and CCR; for HQA, we report Answer EM and Supporting Facts F1; and for SQA and BBH, we report accuracy. All reported results, except the normalized inference cost $T$, are shown as mean $\pm$ standard deviation over three independent runs, while $T$ is reported as the average relative cost across the same three runs with Direct set to 1.0.

The overall trend remains consistent with the main results: SABA achieves the strongest performance across all three DP difficulty levels as well as the general reasoning benchmarks, showing that its gains are not tied to a single backbone. On the most challenging DP-Complex split, SABA attains an SA of 68.8 ($\pm$1.1) and a CCR of 75.7 ($\pm$1.5), outperforming the strongest baselines, where $S^2$R achieves the highest baseline SA of 67.5 ($\pm$1.1) and GoT achieves the highest baseline CCR of 65.1 ($\pm$1.6). SABA also delivers the best results on HQA, with an Ans score of 70.3 ($\pm$0.8) and an SF score of 63.0 ($\pm$0.9), while further achieving the top performance on StrategyQA, with an accuracy of 87.4 ($\pm$0.7), and on BBH, with an accuracy of 86.1 ($\pm$0.9). These results suggest that although the absolute performance of Gemini-1.5-Flash remains below that of DeepSeek-V3, the relative advantage of SABA is preserved across both easy-to-complex detective reasoning and broader reasoning benchmarks, supporting the cross-backbone generality of the framework.

\subsection{Reliability Analysis of QSR-generated Hypotheses}
\label{app:hypothesis_reliability}

\begin{table}[H]
\centering
\footnotesize
\setlength{\tabcolsep}{4pt}
\renewcommand{\arraystretch}{1.1}
\caption{Reliability analysis of QSR-generated hypotheses (\%).}
\label{tab:hypothesis_reliability}
\resizebox{\columnwidth}{!}{%
\begin{tabular}{lccc}
\toprule
Dataset & Unsupported & Flagged & Correction (within flagged) \\
\midrule
DP-Complex & 22 & 27 & 60 \\
HQA        & 24 & 22 & 56 \\
StrategyQA & 31 & 16 & 41 \\
BBH        & 27 & 20 & 49 \\
\bottomrule
\end{tabular}%
}
\end{table}

To further assess the reliability of hypothesis bridging in QSR, we analyze intermediate hypotheses using the per-round logs recorded during inference. For each dataset, we report three statistics: \textit{Unsupported}, the proportion of hypotheses that are not directly supported by evidence; \textit{Flagged}, the proportion of hypotheses marked by the framework as conflicting or weakly supported; and \textit{Correction (within flagged)}, the proportion of flagged hypotheses that are revised or corrected in subsequent iterations. Note that \textit{Unsupported} and \textit{Flagged} may overlap, while \textit{Correction} is computed only within the flagged subset.

As shown in Table~\ref{tab:hypothesis_reliability}, unsupported hypotheses are not rare in QSR, indicating that hypothesis bridging is not risk-free if left unchecked. However, these hypotheses are not propagated to the final conclusion without constraints. On DP-Complex, 22\% of hypotheses are unsupported and 27\% are flagged, while 60\% of the flagged cases are corrected in later iterations; a similar pattern is observed on HQA, where 56\% of flagged hypotheses are corrected. This shows that once risky intermediate bridges are surfaced, later rounds often revise them rather than letting them pass unchanged. On StrategyQA and BBH, the correction rates are lower (41\% and 49\%, respectively), but the framework still identifies and tracks a non-trivial portion of potentially risky hypotheses through explicit flagging. Overall, these results suggest that hypotheses in QSR are not treated as facts, but as auditable intermediate state items that remain subject to IF consistency marking, obstacle-driven reasoning, and per-round log auditing. In this sense, QSR mitigates hallucination risk not by avoiding hypothesis generation altogether, but by keeping unsupported bridges reviewable and partially correctable before final synthesis.

\end{document}